\documentclass[10pt,twocolumn,letterpaper]{article}

\usepackage[final]{cvpr}      

\usepackage{graphicx}
\usepackage{amsmath}
\usepackage{amssymb}
\usepackage{booktabs}
\usepackage{multirow}

\usepackage[pagebackref,breaklinks,colorlinks]{hyperref}

\usepackage[capitalize]{cleveref}
\crefname{section}{Sec.}{Secs.}
\Crefname{section}{Section}{Sections}
\Crefname{table}{Table}{Tables}
\crefname{table}{Tab.}{Tabs.}

\def\cvprPaperID{*****} 
\def\confName{CVPR}
\def\confYear{2022}

\begin{document}

\title{Invariant Meta Learning for Out-of-Distribution Generalization}

\author{Penghao Jiang, Ke Xin, Zifeng Wang, Chunxi Li\\
	The Australian National University, Canberra, Australia * 
}
\maketitle

\begin{abstract}
Modern deep learning techniques have illustrated their
excellent capabilities in many areas, but relies on large
training data. Optimization-based meta-learning train a
model on a variety tasks, such that it can solve new learning
tasks using only a small number of training samples.
However, these methods assumes that training and test data
are identically and independently distributed. To overcome
such limitation, in this paper, we propose invariant meta
learning for out-of-distribution tasks. Specifically, invariant
meta learning find invariant optimal meta-initialization,
and fast adapt to out-of-distribution tasks with regularization
penalty. Extensive experiments demonstrate the effectiveness
of our proposed invariant meta learning on out-ofdistribution
few-shot tasks.
\end{abstract}

\section{Introduction}
\label{sec:intro}
\let\thefootnote\relax\footnotetext{* The first two authors contributed equally as joint first authorship. The
	last two authors contributed equally as joint second authorship.}

Modern deep learning techniques have illustrated their
excellent capabilities in many areas like computer vision,
natural language processing and recommendation, etc \cite{11}.
However, these methods relies on large training data. To
overcome this limitation, few-shot learning methods such
as meta learning has been proposed \cite{6}. Most popular
meta learning approaches is the optimization-based metalearning
\cite{4,17}, which is model-agnostic and can be applied
to various downstream tasks. However, many recent
researches have revealed the vulnerability of machine learning
model when exposed to data with different distributions.

\begin{figure}[!t]
\centering
\includegraphics[width=\columnwidth]{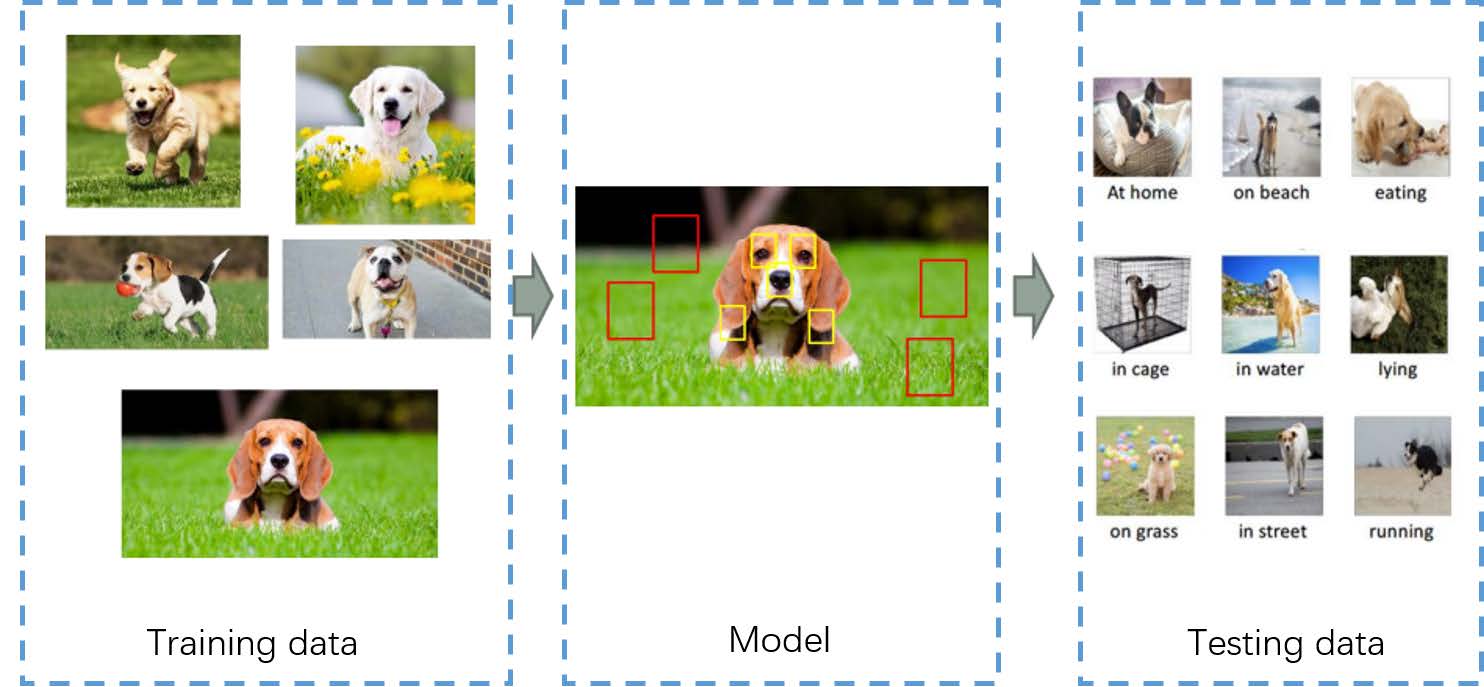}
\caption{Illustration example of how the distribution shifts between
	training data and testing data hamper the performance of
	model predictions.}
\label{fig1}
\end{figure}
\begin{figure}[!t]
	\centering
	\includegraphics[width=\columnwidth]{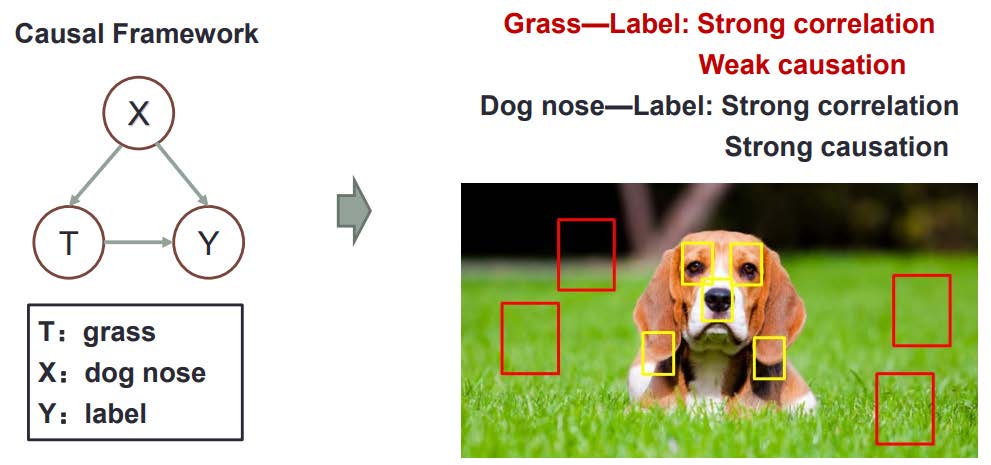}
	\caption{Causal framework of dog perdiction task. Due to the
		spurious correlation, the model tends to focus on both grass and
		dog, which lead to failed prediction in other distributions.}
	\label{fig2}
\end{figure}

Such massive gap is induced by the violation of a fundamental
assumption that training and test data are identically
and independently distributed (a.k.a. i.i.d. assumption),
upon which most of the existing meta learning models are
developed \cite{4,17}. In many real cases where i.i.d. assumption
can hardly be satisfied, especially those high-stake applications
such as healthcare, military and autonomous driving,
instead of generalization within the training distribution,
the ability to generalize under distribution shift is of
more critical significance. As shown in Figure 1, given traning data where dogs are on the grass, model could not make
accurate predictions in testing data where dogs are in water,
cage or street. The reason is that the supurious correlation
between grass and dog in traning data hamper the performance
of model. Due to the spurious correlation, the model
tends to focus on both grass and dog, which lead to failed
prediction in other distribution such as dogs are in water,
cage or street as shown in Figure 2. However, recent meta
learning methods could not overcome the distribution shifts
between training and testing data. In this paper, we consider
a realistic scenario where tasks come from different
distributions (out-of-distribution, OOD).

In this paper, to overcome the problem mentioned
above, we propose Invariant Meta Learning (IML) for out-of-
distribution tasks, a general learning framework that jointly adjusts gradient magnitudes and directions. Specifically,
invariant meta learning find invariant optimal metainitialization,
and fast adapt to out-of-distribution tasks with
regularization penalty. To summarize, our main contributions
are:
\begin{itemize}
\item We consider the challenge of out-of-distribution tasks
faced by few-shot learning, we show a natural idea to
jointly adjust gradient magnitudes and directions of all
tasks in the meta optimization process;
\item We propose Invariant Meta Learning (IML) for out-ofdistribution
tasks, a general learning framework that
jointly adjusts gradient magnitudes and directions;
\item We conduct extensive experiments and analysis to
demonstrate that our approach effectively improves the
performance and generalization ability under both indistribution
and out-of-distribution few-shot settings,
and thus it can be regarded as a better baseline.
\end{itemize}

\section{Method}
In this section, we introduce our proposed Invariant Meta
Learning (IML) to address the out-of-distribution problem
in few-shot tasks. IML learns invariant optimal predictors
based on optimization based meta learning framework.
To learn invariant optimal meta-initialization in optimization
based meta learning, the main challenge is that OOD
problem exacerbates the inconsistency in both task-gradient
magnitudes and directions. To overcome such problem,
IML finds invariant optimal initialization, and adapt to outof-
distribution tasks with regularization penalty.

\textbf{Model-agnostic meta-learning (MAML)} \cite{4} is an approach to optimization-based meta-learning that is related to our work. For some parametric model $f_\theta$, MAML aims to find a single set of parameters $\theta$ which, using a few optimization steps, can be successfully adapted to any novel task sampled from the same distribution. For a particular task instance $\mathcal{T}_i=\left(\mathcal{D}^{t r}, \mathcal{D}^{v a l}\right)$, the parameters are adapted to task-specific model parameters $\theta_i^{\prime}$ by applying some differentiable function, typically an update rule of the form:
\begin{equation}
\theta_i^{\prime}=\mathcal{G}\left(\theta, \mathcal{D}^{t r}\right),\label{eq1}
\end{equation}
where $\mathcal{G}$ is typically implemented as a step of gradient descent on the few-shot training set $\mathcal{D}^{\text {tr }}, \theta_i^{\prime}=\theta-$ $\alpha \nabla_\theta \mathcal{L}_{\mathcal{T}_i}^{t r}\left(f_\theta\right)$. Generally, multiple sequential adaptation steps can be applied. The learning rate $\alpha$ can also be metalearned concurrently, in which case we refer to this algorithm as Meta-SGD \cite{13}. During meta-training, the parameters $\theta$ are updated by back-propagating through the adaptation procedure, in order to reduce errors on the validation set $\mathcal{D}^{ {val }}$ :
\begin{equation}
\theta \leftarrow \theta-\eta \nabla_\theta \sum_{\mathcal{T}_i \sim p(\mathcal{T})} \mathcal{L}_{\mathcal{T}_i}^{v a l}\left(f_{\theta_i^{\prime}}\right).\label{eq2}
\end{equation}

The approach includes the main ingredients of optimization-based meta-learning with neural networks: initialization is done by maintaining an explicit set of model parameters $\theta$; the adaptation procedure, or ``inner loop", takes $\theta$ as input and returns $\theta_i^{\prime}$ adapted specifically for task instance $\mathcal{T}_i$, by iteratively using gradient descent (Eq. \ref{eq1}); and \textit{termination}, which is handled simply by choosing a fixed number of optimization steps in the ``inner loop". MAML updates $\theta$ by differentiating through the ``inner loop" in order to minimize errors of instance-specific adapted models $f_{\theta_i^{\prime}}$ on the corresponding validation set (Eq. \ref{eq2}). We refer to this process as the ``outer loop" of meta-learning. We use the same stages to describe IML.

\textbf{Invariant Meta Learning (IML)} finds invariant optimal
meta-initialization, and fast adapt to out-of-distribution
tasks with regularization penalty. MAML fast adapt network
to new task during the inner loop and learns universal
meta-initialization in outer loop. Similarly, in IML, we
update network with the bi-level update, optimizing classifier
in the inner loop and learning feature representation in the outer loop. For the inner-level optimization, the parameters $\theta$ of the predictor become $\theta_i$ while adapting to the task $t_i \in \mathcal{T}_{t r}$. This correspond to the inner optimization of MAML, except that each task $t_i$ has a corresponding network $\theta_i$. The optimization in the inner loop can be defined as follows:
\begin{equation}
\theta_i^{\prime}=\theta-\alpha \nabla_\theta \mathcal{L}_{\mathcal{T}_i}^{t r}\left(f_\theta\right)
\end{equation}
where $\alpha$ is a learning rate of the inner optimization.

With inner optimized network $f_{\theta_i^{\prime}}$, we have outer loop objective function with variance penalty regularizer:
\begin{equation}
\mathcal{L}^{ {val }}=\sum_{\mathcal{T}_i \sim p\left(\mathcal{T}^{t r}\right)} \sum_{\mathcal{T}_j \sim p\left(\mathcal{T}^{ {val }}\right)} \mathcal{L}_{\mathcal{T}_j}^{ {val }}\left(f_{\theta_i^{\prime}}\right)
\end{equation}
\begin{equation}
\theta \leftarrow \theta-\eta \nabla_\theta \mathcal{L}^{ {val }}-\beta \lambda \operatorname{trace}\left(\operatorname{Var}_{\mathcal{T}^{ {val }}}\left(\nabla_\theta \mathcal{L}^{ {val }}\right)\right)
\end{equation}
where $\eta, \beta$ are the learning rate of the outer loop optimization, $t_j$ is task $j$ for outer loop optimization for the network $\theta_i^{\prime}, \mathcal{L}$ is the loss function for outer loop optimization. Note that the inner optimized network $f_{\theta_i^{\prime}}$ is used to update meta-initialization in outer loop with $t_j$ whereas it is updated from meta-initialization with $t_i$ in ther inner loop. IML learn invariant meta-initialization obtained from the discrepancy among different training tasks with variance penalty regularizer.

\begin{table*}
\centering
\begin{tabular}{ccccccc}
	\hline
	\multirow{2}{*}{Method}&\multicolumn{2}{c}{miniImageNet} & \multicolumn{2}{c}{CUB}&\multicolumn{2}{c}{SUN} \\\cline{2-7}
	& 5-way 1-shot& 5-way 5-shot& 5-way 1-shot& 5-way 5-shot& 5-way 1-shot& 5-way 5-shot\\\hline
	 Meta-Learner LSTM & 24.99 & 29.79 & 36.23 & 44.39 & 30.99 & 44.86 \\
	MAML & 45.69 & 60.90 & 48.87 & 63.99 & 57.75 & 71.45 \\
	Reptile & 26.59 & 39.87 & 27.21 & 42.35 & 28.30 & 51.62 \\
	Matching Network & 47.63 & 56.28 & 53.06 & 62.19 & 55.02 & 62.57 \\
	Prototypical Network & 46.15 & 65.56 & 48.21 & 57.80 & 55.70 & 67.32 \\
	Relation Network & 47.64 & 63.65 & 52.76 & 64.71 & 58.29 & 72.15 \\
	Baseline & 23.84 & 32.09 & 25.14 & 35.35 & 27.44 & 34.54 \\
	Baseline++ & 30.15 & 41.19 & 32.48 & 42.43 & 35.56 & 44.42 \\
	\textbf{IML} & \textbf{48.35} & \textbf{67.21} & \textbf{54.18} & \textbf{65.85} & \textbf{59.24} & \textbf{74.18} \\
	\hline
\end{tabular}
\caption{Average accuracy (\%) comparison to state-of-the-arts with 95\% confidence intervals on 5-way classification tasks under the
	in-distribution FSL setting. Best results are displayed in boldface.}

\end{table*}

\begin{table*}
	\centering
	\begin{tabular}{ccccccc}
		\hline
		\multirow{2}{*}{Method} &\multicolumn{2}{c}{miniImageNet$\rightarrow$ CUB}&\multicolumn{2}{c}{miniImageNet$\rightarrow$ SUN}&\multicolumn{2}{c}{CUB$\rightarrow$miniImageNet} \\\cline{2-7}
		& 5-way 1-shot& 5-way 5-shot& 5-way 1-shot& 5-way 5-shot& 5-way 1-shot& 5-way 5-shot\\\hline
			 \text { Meta-Learner LSTM } & 23.77 & 30.58 & 25.52 & 32.14 & 22.58 & 28.18 \\
			\text { MAML } & 40.29 & 53.01 & 46.07 & 59.08 & 33.36 & 41.58 \\
			\text { Reptile } & 24.66 & 40.86 & 32.15 & 50.38 & 24.56 & 40.60 \\
			\text { Matching Network } & 38.34 & 47.64 & 39.58 & 53.20 & 26.23 & 32.90 \\
			\text { Prototypical Network } & 36.60 & 54.36 & 46.31 & 66.21 & 29.22 & 38.73 \\
			\text { Relation Network } & 39.33 & 50.64 & 44.55 & 61.45 & 28.64 & 38.01 \\
			\text { Baseline } & 24.16 & 32.73 & 25.49 & 37.15 & 22.98 & 28.41 \\
			\text { Baseline++ } & 29.40 & 40.48 & 30.44 & 41.71 & 23.41 & 25.82 \\
			{ \textbf{IML} } & \textbf{41.27} & \textbf{57.34} & \textbf{50.42} & \textbf{69.15} & \textbf{34.26} & \textbf{44.17}\\
		\hline
	\end{tabular}
	\caption{Average accuracy (\%) comparison to state-of-the-arts with 95\% confidence intervals on 5-way classification tasks under the
		in-distribution FSL setting. Best results are displayed in boldface.}
	
\end{table*}

\section{Experiments}
\paragraph{Datasets.}
In this paper, we address the few-shot classification
problem under both in-distribution and out-ofdistribution
FSL settings. These settings are conducted on three benchmark datasets: miniImageNet \cite{24}, Caltech-
UCSD-Birds 200-2011 (CUB) \cite{26}, and SUN Attribute
Database (SUN) \cite{16}.

\paragraph{Baselines.} To evaluate the effectiveness of the proposed
framework, we consider the following representative meta
learning methods on the few-shot image classification task:
MAML \cite{5}, Reptile \cite{14}, Matching Network \cite{24}, Prototypical
Network \cite{21}, Relation Network \cite{22}, Baseline and
Baseline++ \cite{3}.

\paragraph{Experimental Settings.} We conduct experiments on 5-way 1-shot and 5-way 5 -shot settings, there are 15 query samples per class in each task. We report the average accuracy (\%) and the corresponding $95 \%$ confidence interval over the 2000 tasks randomly sampled from novel classes. To fairly evaluate the original performance of each method, we use the same 4-layer ConvNet \cite{24} as the backbone for all methods and do not adopt any data augmentation during training. All methods are trained via SGD with Adam \cite{10}, and the initial learning rate is set to $e^{-3}$. For each method, models are trained for 40,000 tasks at most, and the best model on the validation classes is used to evaluate the final reporting performance in the meta-test phase.

\paragraph{Evaluation Using the In-Distribution Setting.} Table 1
shows the comparative results under the in-distribution FSL
setting on three benchmark datasets. It is observed that IML outperforms the original MAML in all in-distribution FSL scenarios. For 1-shot and 5-shot on miniImageNet $\rightarrow$ miniImageNet, IML achieves about $1 \%$ higher performance than Prototypical Network. However, IML achieves $5 \%$ and $10 \%$ higher performance for 1-shot and 5-shot on CUB $\rightarrow$ CUB, and $3 \%$ and $6 \%$ higher performance on SUN $\rightarrow$ SUN. As the latter two scenarios are conducted on finegrained classification datasets, we attribute the promising improvement to that the categories in these fine-grained datasets share more local concepts than those in coarsegrained datasets, and thus a more discriminative space can be rapidly learned with a few steps of adaptation. Moreover, IML achieves the best performance among all baselines in all in-distribution FSL scenarios, which shows that our approach can be considered as a better baseline option under the in-distribution FSL setting.

\paragraph{Evaluation Using the Out-of-Distribution Setting.} We
also conduct out-of-distribution FSL experiments and report
the comparative results in Table 2. Compared to the results
under the in-distribution setting, it can be observed that
all approaches suffer from a larger discrepancy between the
distributions of training and testing tasks, which results in
a performance decline in all scenarios. However, IML still
outperforms the original MAML in all out-of-distribution
FSL scenarios, demonstrating that the bilevel optimization  strategy for adaptation and the learning of transferable latent
factors can be utilized to improve simple meta learning
approaches. Also, IML achieves all the best results, indicating
that our approach can be regarded as a promising baseline
under the out-of-distribution setting.

\section{Conclusion}

In this paper, we consider the challenge of out-ofdistribution
tasks faced by few-shot learning. We propose
Invariant Meta Learning (IML) for out-of-distribution tasks,
a general learning framework that jointly adjusts gradient
magnitudes and directions. Extensive experiments demonstrate
that our approach effectively improves the performance
and generalization ability under both in-distribution
and out-of-distribution few-shot settings, and thus it can be
regarded as a better baseline.

\nocite{*}
{\small
\bibliographystyle{ieee_fullname}
\bibliography{egbib}
}

\end{document}